# FEEL: A Framework for Evaluating Emotional Support Capability with Large Language Models

Huaiwen Zhang[*], Yu Chen[*], Ming Wang and Shi Feng[(✉)]

School of Computer Science and Engineering, Northeastern University, Shenyang 110819, China
`fengshi@cse.neu.edu.cn`

**Abstract.** Emotional Support Conversation (ESC) is a typical dialogue that can effectively assist the user in mitigating emotional pressures. However, owing to the inherent subjectivity involved in analyzing emotions, current non-artificial methodologies face challenges in effectively appraising the emotional support capability. These metrics exhibit a low correlation with human judgments. Concurrently, manual evaluation methods extremely will cause high costs. To solve these problems, we propose a novel model FEEL (Framework for Evaluating Emotional Support Capability with Large Language Models), employing Large Language Models (LLMs) as evaluators to assess emotional support capabilities. The model meticulously considers various evaluative aspects of ESC to apply a more comprehensive and accurate evaluation method for ESC. Additionally, it employs self-CoT and probability distribution approaches for a more stable result and integrates an ensemble learning strategy, leveraging multiple LLMs with assigned weights to enhance evaluation accuracy. To appraise the performance of FEEL, we conduct extensive experiments on existing ESC model dialogues. Experimental results demonstrate our model exhibits a substantial enhancement in alignment with human evaluations compared to the baselines. Our source code is available at https://github.com/Ansisy/FEEL.

## 1 Introduction

Emotional Support Conversations (ESC) [1] is a goal-oriented task focused on alleviating emotional distress and effecting positive changes in individuals' psychological states. ESC has extensive applications in mental health support, customer service chats, etc. However, the evaluation of ESC, as compared to the generation, presents a more challenging task. As shown in Fig. 1, unreliable ESC evaluation systems may mislead users and even increase their psychological stress. In the available studies, works on ESC evaluation are mainly divided into two categories. The first is to use traditional automatic metrics, such as BLEU [2], ROUGE [3], Distinct-n [4],

---

[*] These authors contributed equally to this work.

METEOR [5], etc. These metrics predominantly concentrate on the similarity to the golden outputs. They exhibit a low correlation with human evaluations. Consequently, these methods are not suitable for assessing ESC as they cannot understand and evaluate complex and diverse emotions of humanity. The second strategy involves training individuals to assess dialogue quality in specific aspects, necessitating annotators with robust text evaluation skills. This methodology has demonstrated efficacy in assessing the quality of ESC. However, there is no systematic evaluation framework and the basis for evaluation relies heavily on the subjective experience of the authors. This variation necessitates humans to be proficient in the meaning of different aspects which significantly increases labor costs. Consequently, this approach demands substantial time and labor investment. Furthermore, manual evaluations typically involve only a limited subset of data samples. If the sampling methodology is not reasonably chosen, it may introduce bias into the results.

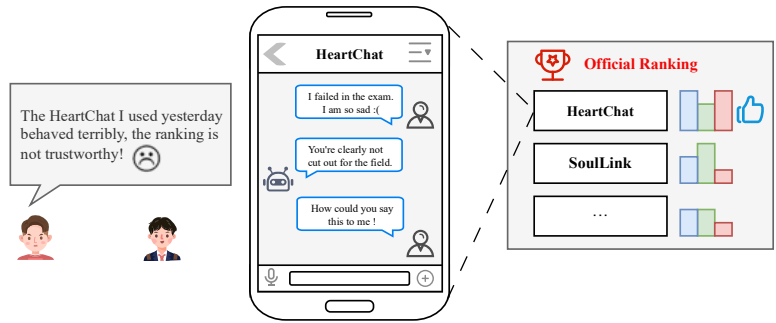

**Fig. 1.** Low-quality evaluation systems lead to poor emotional support experiences and even worsen user's situation.

Recently, many evaluation methods for natural language generation (NLG) based on LLMs have achieved notable advancements. Fu *et al.* [6] discovered that generative pre-training models exhibit enhanced reliability when guided by specific task and aspect definitions, thereby laying the groundwork for the flexible selection of aspects. Moreover, Liu *et al.* [7] introduced an LLM-based evaluator for text summary assessment and achieved good alignment with humans. These works unveil the potential of LLMs as evaluators in assessing task-specific dialogues. Nonetheless, due to the stochastic nature of computations in the underlying hardware, the responses generated by a single LLM to the same question tend to be unstable with significant variance [8]. Furthermore, most current works focus on generic evaluations of NLG tasks, with a notable lack of specialized prompts designed for the assessment of ESC. Therefore, to get results with better human alignment, a more stable and task-specific mixed LLMs model is needed for the evaluation task of ESC.

To tackle these issues, we first redefine six evaluation aspects in emotional support skills and text quality dimensions shown in AUGESC [9] by analyzing interactions within psychotherapy talk [10] to systematically evaluate the emotional support capa-bility of dialogue systems. Furthermore, we propose a novel evaluation framework FEEL, a.k.a. Framework for Evaluating Emotional support capability with Large Language Models. Utilizing an ensemble learning method, FEEL integrates three LLMs: ERNIE-Bot 4.0 [11], GLM-4 [12], and GPT-3.5-Turbo [13]. By providing LLMs with task definitions and scoring criteria as prompts, we empower the LLMs to repeatedly derive the score distribution probability for each aspect. Moreover, we average the scores to mitigate the effects of variance. Then, we build an emotional support capability score dataset ESCEval by instructing annotators to meticulously assess with dialogues in AUGESC [9] and ESConv [1]. Subsequently, we use the LLMs' Spearman's rank correlation coefficient with the human result in ESCEval as weights in the framework to derive the final FEEL outputs: the score of emotional support capabilities. Finally, we conduct extensive experiments on dialogues generated from existing emotional support models. In comparison to automatic evaluation metrics, FEEL demonstrates superior alignment with human-derived outcomes.

The main contributions of this paper are summarized as follows:

- We refine a suite of aspects, which assess dialogue quality in terms of both emotional support skills and text quality.
- We propose a framework FEEL for emotional support capability evaluation based on mixed LLMs.
- We annotate ESCEval: a high-quality dataset composed of human evaluations of emotional support capabilities, using instances from ESConv and AUGESC and train the weights of LLMs on the dataset.
- We carry out extensive experiments on different emotion support models, demonstrating that FEEL surpasses existing evaluation metrics in aligning with human evaluation.

## 2   Related Work

**Non-LLM-based ESC.** In terms of datasets, Liu *et al*. [1] created an Emotion Support Conversation dataset, richly annotated for both help-seeker and supporter interactions. Meanwhile, Zheng *et al*. [9] developed AUGESC, an augmented dataset for ESC tasks. In terms of emotional support modeling, Peng *et al*. [14] proposed a Feedback Aware Double Controlling Network for strategic scheduling and supportive response generation. Peng *et al*. [15] introduced a Global-to-Local Hierarchical Graph Network to capture multi-source information and model hierarchical conversation relationships. Tu *et al*. [16] propose the novel MISC model, which first infers users' fine-grained emotional states, then skillfully responds with a strategic mix. To further improve the effectiveness of ESC, Zhao *et al.* [17] suggest considering turn-level state transitions in ESC, encompassing semantic, strategic, and emotional shifts. But all the above models, which give limited help to human emotional support tasks, are not very effective.

**LLM-based ESC.** To test LLM's ability to perform on ESC tasks, Li *et al*. [18] conducted experiments on 45 ESC tasks with various LLMs, demonstrating their grasp of emotional intelligence. Zhang *et al*. [19] reported that although LLMs perform satisfactorily in simpler tasks, they fall short in complex tasks requiring deep understanding or structured sentiment analysis. It is shown that ESC tasks using the LLM performed better on overall emotional support than previous models.

### 2.1 NLG Evaluation Models

**Non-LLM-based Evaluation Models.** To address the poor performance of traditional automatic evaluation metrics on some NLG tasks, Tao *et al*. [20] introduced RUBER, blending referenced and unreferenced metrics for evaluating replies against both ground truth and queries. Zhong *et al*. [21] developed UNIEVAL, a unified evaluator for NLG across multiple dimensions. Further, Mehri *et al.* [22] presented USR, a reference-free metric using unsupervised models to assess dialog qualities. These models show good performance on text quality evaluation, but the evaluation aspects of ESC evaluation tasks are quite different from those of traditional text quality evaluation tasks, so it is difficult to apply these models directly to ESC evaluation tasks.

**LLM-based Evaluation Models.** To test whether LLM has the ability to evaluate the quality of NLG tasks, Fu *et al*. [6] introduced GPTSCORE, an innovative framework leveraging generative models for text assessment. Liu *et al*. [7] developed G-EVAL, utilizing large models with CoT and form-filling to evaluate NLG quality. To further energize the LLM in evaluating the quality of NLG tasks, Liu *et al.* [23] proposed AUTOCALIBRATE, a method for aligning LLM-based evaluations with human preferences without gradients. Chen *et al*. [24] examined three reference-free evaluation techniques using ChatGPT or similar LLMs for dialog generation tasks. The same problem as non-LLM-based evaluation models, these models are difficult to apply to the ESC evaluation task.

## 3 Methodology

### 3.1 Task Definition

To evaluate the emotional support capability, it is imperative to consider both the dialogue generation quality and the emotional perception in ESC. Currently, most ideas of ESC evaluation aspects selection follow the procedure of emotional support—exploration, comforting, and action—as outlined in ESConv [1]; however, the specific selections demonstrate considerable variability and a notable lack of systematic approach across different studies. Therefore, we need to first make a comprehensive and systematic definition of ESC evaluation.

Building upon the methodology proposed by AUGESC [9], we categorize the evaluation of ESC into two distinct dimensions: emotional support skills and text quality. Subsequently, through research on psychotherapy in [10], we further explain each aspect as below:

**Emotional Support Skills.** (1) **Informativeness**: the extent to which the supporter guides the help-seeker in articulating their emotional problems in detail. (2) **Comprehensibility**: the extent to which the supporter understands the seeker's experiences and feelings. (3) **Helpfulness**: the supporter's capacity to alleviate the client's emotional distress and provide constructive advice.

**Text Quality.** (1) **Consistency**: the alignment of the supporter's perspective and the adherence to his role. (2) **Coherence**: the maintenance of focus on the subject matter and the fluidity of topic transitions. (3) **Safety**: the absence of inappropriate language or content within the dialogue.

For each aspect, a four-tiered Likert scale score $S_i$ is assigned, where $S_i \in [0,3]$ and can be either an integer or a decimal. The evaluation result of one dialogue is represented as $\{S_1, S_2, S_3, S_4, S_5, S_6\}$, denoting the emotional support capability scores across the six aspects.

The complete evaluation criteria for human annotators is shown on our open-source website.

### 3.2 ESCEval: A Dataset of Human ESC Evaluation

In order to assess the emotional support capabilities of each LLM and then to determine their weight in FEEL, it is imperative to establish a dataset comprising human scores for emotional support dialogues as a reference standard for LLM. Therefore, we construct ESCEval: a dataset composed of human evaluations of emotional support capabilities. We enlisted six college students majoring in computer science to evaluate a total of 200 dialogue instances, randomly selected from ESConV and AUGESC. Any identifiable personal information present in the dialogues is anonymized to ensure data security. Subsequently, refer to the human annotations work in [25], we conduct a total of two scoring rounds: in the first round, each annotator is asked to score in six aspects as per the standards delineated in Sect. 3.1; in the second round, we manually inspect the results and require that each annotator's score on a dimension differs from another annotator by more than 1 point, and that the other annotators' scores within 1 point of each other are rescored. Whenever a scoring discrepancy greater than one point arose without the pattern, all annotators were instructed to collaboratively reevaluate the annotation. When rescoring, the annotator can refer to other people's scores to review again to avoid strong subjectivity. We finally calculate the average score of all annotators to construct the dataset. By employing this methodical approach, we effectively reduce the impact of personal subjectivity on the scoring results and construct a high-quality emotional support capability scores dataset. The process of ESCEval construction can be seen in Fig. 2.

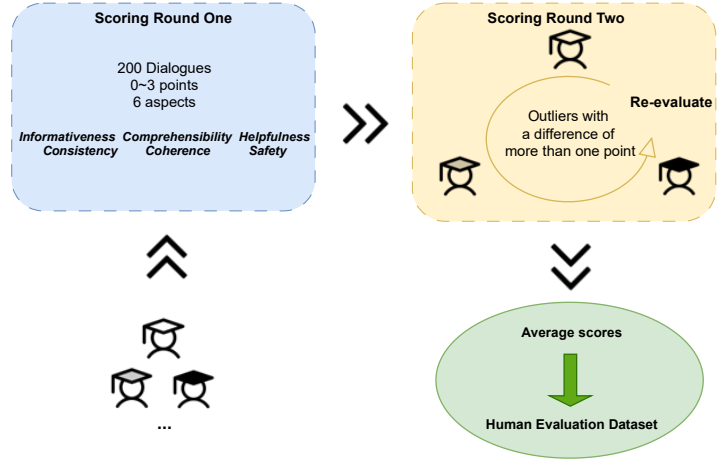

**Fig. 2**. The process of ESCEval construction

### 3.3 Proposed FEEL

FEEL represents a comprehensive, multifaceted evaluator based on LLMs, comprising: a) a structured prompt delineating task specification; evaluative criteria for each aspect as discussed in Sect. 3.1; the evaluation sample dialogue; and the output format that standardize the output form of LLMs; b) an advanced scoring algorithm that ascertains individual LLM scores by computing probabilities linked to the assignment of particular scoring bands to get stable evaluation scores; and c) a comprehensive integrative weighted computational approach using ensemble learning method that synthesizes results from multiple LLMs to deduce definitive evaluation scores. The output of the FEEL metric quantitatively reflects the emotional support capabilities of the supporter exhibited in the sample across each aspect. The detailed framework of FEEL is shown in Fig. 3.

**Evaluator Prompt Design.** A prompt serves as a natural language directive that clearly delineates the specific evaluation criteria and expectations for an assessment task. We first delineate prompt for the task of evaluating emotional support capabilities:

> *You will play the role of a psychologist who is well versed in emotional support. There will be a dialog between the help seeker (i.e., the person seeking support) and the supporter (i.e., the person providing support)………*

Furthermore, the prompt phrases encompass specific evaluation aspects pertinent to the task of assessing emotional support capability:

> *Evaluation Criteria:*
> *Comprehensibility: how well the supporter understands the help-seeker's experiences and feelings……*

The complete prompt is shown in our open-source website.

**Self Chain-of-Thoughts for Emotional Capability Evaluation.** Chain-of-Thought (CoT) prompting in the context of large language models refers to a technique that enhances the model's reasoning capabilities by breaking down complex tasks into a sequence of intermediate steps. In the emotional support capability evaluation, the LLMs needs to master the detailed evaluation steps of each indicator so as to conduct detailed evaluation step by step. In the meanwhile, manually designing CoT is time-consuming and cannot accurately follow the problem-solving logic of LLM. Therefore, refer to the CoT design in [7], we give LLM the task definition and the specific metric evaluation criteria to let it generate the evaluation steps automatically (Self-CoT). For example, for evaluating informativeness in emotional support capability, we add a line of "Evaluation Steps:" to the prompt and let LLM to generate the following CoT automatically.

> *1. Read the client's description carefully.*
> *2. Extract the emotional issues and related details in the description.*
> *3. Evaluate whether the description contains the specific causes of the emotional issues.*
> *……*

**Individual LLM Evaluation.** For specific evaluation aspect of single-round dialog data, we utilize a prompt-based answer format to facilitate the LLM in generating selection probabilities for each score band (0 to 3), respectively, with the probabilities of the four score bands summing to 1:

> *Answer format (give the probability of each score band for each type of score):*
> *- Comprehensibility Score:*
> *0 points:*
> *1 point:*
> *2 points:*
> *3 points:*

Subsequently, for a particular evaluation aspect, the selection probabilities produced by the LLM serve as weights. These are multiplied by the scores within the respective score bands, adding up the weighted scores of the four score bands to ascertain the singular-round score for the specific aspect. Acknowledging the inherent volatility in LLM responses to ascertain the singular-round score for the specific aspect, the variability of scoring outcomes across different LLM rounds [26]. For the same evaluation aspect of the same round of dialog, we mitigate it by calculating the average of ten successive LLM round scores, thereby deriving a stable final score:

$$S_i = \frac{\sum_{n=1}^{10}\sum_{j=0}^{3} W_{j,n} * j}{10} \quad (1)$$

where $S_i$ represents individual LLM score in aspect $i$, $W_{j,n}$ denotes the selection probability of score band $j$ in iteration $n$ ($j$ is an integer ranging from 0 to 3 and $n$ is an integer ranging from 1 to 10).

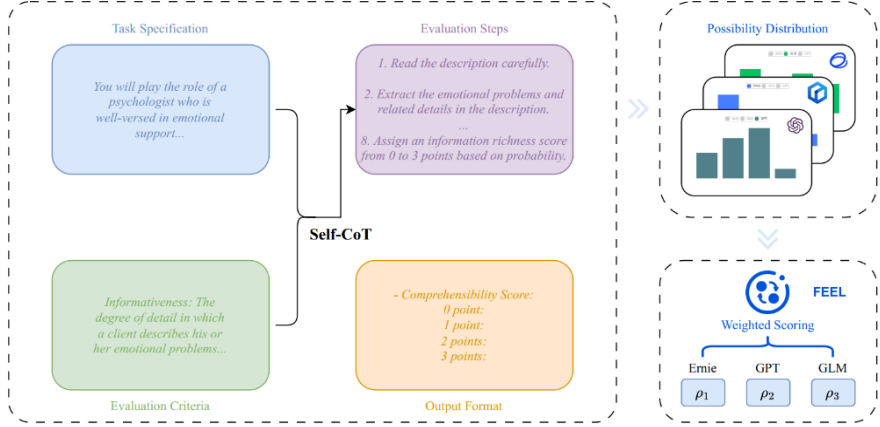

**Fig. 3.** The detailed framework of FEEL. We first input Task Specification and Evaluation Criteria to each LLM and ask it to generate a self-CoT of detailed Evaluation Steps. Then we employ the LLMs to output the possibility distribution according to the Output Format for every dialogue sample. Finally, we use the Spearman Correlation-weighted score by three LLMs as the FEEL output.

**Establishment of FEEL.** Following individual LLM evaluations, as diverse LLMs demonstrate distinct advantages across different dialogues, allocating the scores of each model through weight adjustment can prominently underscore the individual strengths of each model, thereby enhancing the overall evaluation outcomes. After preliminary testing, we identify three LLMs: ERINE-Bot-4.0 (LLM-1), GLM-4 (LLM-2), and GPT-3.5-turbo (LLM-3) as the constituents of FEEL, which all have high-performance natural language understanding and fine-grained emotional analysis capabilities. Subsequently, to determine the weight of each LLM in FEEL, we respectively employ the three LLMs to evaluate the ESCEval in Sect. 3.2. Finally, using the Spearman's rank correlation coefficient with results in the ESCEval as the weight, we calculate the FEEL result as follows:

$$\rho_{i,n} = \frac{c_{i,n}}{\sum_{n=1}^{3} c_{i,n}} \quad (2)$$

$$F_i = \sum_{n=1}^{3} \rho_{i,n} * S_{i,n} \qquad (3)$$

where $\rho_{i,n}$ denotes the weight of LLM-n in aspect $i$, $C_{i,n}$ represents the Spearman correlation coefficient of LLM-n in aspect $i$. $S_{i,n}$ represents the individual score of LLM-n in aspect $i$ and $F_i$ denotes the FEEL score in aspect $i$, which represents the dialogue's emotional support capability in aspect $i$.

## 4    Experiments and Results

### 4.1    Implementation of FEEL

**Experimental Settings.** We utilize the APIs of each LLM in FEEL and send request over the network to the model hosting server to access and interact with the corresponding LLMs for subsequent experimentation. The whole experiment was completed on 2024-02-26.

**Evaluation Implementation of Individual LLM.** For each of the three LLMs: GPT-3.5-turbo, ERNIE-bot-4.0, and GLM-4, we execute rigorous performance evaluations. Utilizing the standardized prompts delineated previously, we train and assess 200 dialogue instances collected in Sect. 3.2. Across various dialogue datasets, we ascertain the performance of these LLMs across six evaluative aspects. Subsequently, we conduct an analysis of the Spearman correlation coefficients and Kendall's tau coefficient, comparing the evaluation scores of these LLMs across the six aspects against ESCEval. The calculation formulas are as follows:

$$r = 1 - \frac{6\sum_i^n d_i^2}{n(n^2-1)} \qquad (4)$$

where $r$ is the Spearman rank correlation coefficient, $n$ is the number of data points, $i$ denotes that this is the $i$th data, and $d_i$ denotes the gap between the model's evaluation score and the manual labeling score.

$$\tau = \frac{C-D}{\sqrt{(C+D+T)(C+D+U)}} \qquad (5)$$

where $\tau$ is the Kendall's tau coefficient. $C$ is the number of consistent pairs. $D$ is the number of inconsistent pairs. $T$ is the number of leveled pairs for the first variable. $U$ is the number of leveled pairs for the second variable.

**Weight Determination and Analysis.** Following the individual LLM evaluation and the establishment of FEEL in Sect. 3.3, we employ the three LLMs' Spearman rank correlation coefficient on ESCEval to derive the final FEEL score using equation 3, the pertinent experimental result are detailed in Table 1 and Table 2. A substantial number of the coefficient values more than 0.3 and some approached or exceeded 0.4, indicating that the three LLMs demonstrated a high correlation with human assess-ments across various aspects, and FEEL effectively synthesizes the high-performance aspects of each LLM using the calculation formula, reaching the highest Spearman's rank correlation coefficient of 0.509 in terms of helpfulness. The FEEL result is shown in the last lines of two tables.

It should be emphasized that while FEEL generally outperforms individual LLMs across a multitude of aspects, its efficacy might be compromised on certain specific aspects due to the pronounced variability inherent in the dataset or a potential predisposition towards a particular model.

**Table 1.** Spearman's rank correlation coefficient (Spear.) and Kendall's tau coefficient (Kend.) for different models on the aspects of emotional support skills.

|  | Informativeness | | Comprehensibility | | Helpfulness | |
| --- | --- | --- | --- | --- | --- | --- |
|  | Spear. | Kend. | Spear. | Kend. | Spear. | Kend. |
| ERNIE-4.0 | 0.368 | 0.270 | 0.372 | 0.269 | 0.414 | 0.313 |
| GPT-3.5 | 0.163 | 0.119 | 0.190 | 0.139 | 0.360 | 0.264 |
| GLM-4 | 0.364 | 0.299 | 0.317 | 0.250 | 0.385 | 0.297 |
| **FEEL** | **0.404** | **0.300** | **0.429** | **0.314** | **0.509** | **0.377** |

**Table 2.** Spearman's rank correlation coefficient (Spear.) and Kendall's tau coefficient (Kend.) for different models on the aspects of text quality.

|  | Consistency | | Coherence | | Safety | |
| --- | --- | --- | --- | --- | --- | --- |
|  | Spear. | Kend. | Spear. | Kend. | Spear. | Kend. |
| ERNIE-4.0 | 0.427 | 0.323 | **0.343** | **0.250** | 0.384 | 0.298 |
| GPT-3.5 | 0.126 | 0.088 | 0.135 | 0.094 | 0.257 | 0.192 |
| GLM-4 | 0.313 | 0.244 | 0.265 | 0.211 | 0.311 | 0.252 |
| **FEEL** | **0.434** | **0.327** | 0.331 | 0.241 | **0.409** | **0.314** |

### 4.2 Comparative Results

**Sample Construction.** To comprehensively ascertain FEEL's advancement in evaluating the emotional support capabilities of models, we select three pre-trained models: BlenderBot-Joint [1], MISC [16], TransESC [17] and four large language models: Spark-V3.0 [27], Baichuan2-Turbo [28], qwen-turbo [29], ChatGLM-6B [30]. Ten distinct conversation topics are selected, and the reply lengths are standardized across different models within a specified range to control irrelevant variables. Subsequently, based on the methodology in [17], the human evaluators are required to choose which one performs better (or tie) in every two models following five aspects (1) Fluency, (2) Identification, (3) Empathy, (4) Suggestion and (5) Security. Finally, for each topic, the tally of wins versus losses for each model is computed to establish a model ranking of emotional support capabilities as a standard of comparison for subsequent evaluations.

**Baselines.** We compare FEEL with the following automatic metrics which are widely used in ESC evaluation:

- BLEU-1, BLEU-2 [2] measure the n-gram precision between the generated text and reference texts, specifically focusing on unigrams and bigrams, respectively.
- ROUGE-1, ROUGE-2 and ROUGE-L [3] measure the lexical overlap between the generated text and corresponding references based on unigram, bigram and longest common subsequence, respectively.
- Meteor [4] measures the alignment between the generated text and reference texts by considering exact, stem, synonym, and paraphrase matches to compute a harmonic mean of precision and recall.

**Evaluation Strategy.** To assess the extent to which FEEL correlates with human judgment more effectively than traditional automatic metrics, we use the rank-based indicators in Sect. 4.1: (1) Spearman's rank correlation coefficient; (2) Kendall's Tau. Concurrently, to quantify the discrepancies in error between the model predictions and human rankings across the sample, we also utilize (3) Root mean squared error and (4) Mean absolute error as our metrics:

$$RMSE = \frac{\sqrt{\sum_{i=1}^{n}(p_i - r_i)^2}}{n} \tag{6}$$

$$MAE = \frac{\sum_{i=1}^{n}|p_i - r_i|}{n} \tag{7}$$

where $p_i$ represents the model prediction ranking, $r_i$ represents the manual ranking result, and $n$ is the number of models participating in the ranking. The comparative calculation results are shown in Table 3. It can be seen that FEEL is significantly better than all other baselines in four metrics.

**Table 3.** The average Spearman's rank correlation coefficient (Spear.), Kendall's Tau (Kend.), Rooted Mean Squared Error (RMSE) and Mean Absolute Error (MAE) on the sample.

|         | Spear. | Kend.  | RMSE  | MAE   |
|---------|--------|--------|-------|-------|
| BLEU-1  | -0.136 | -0.124 | 2.868 | 2.400 |
| BLEU-2  | -0.082 | -0.076 | 2.878 | 2.343 |
| ROUGE-1 | -0.261 | -0.210 | 3.145 | 2.714 |
| ROUGE-2 | -0.332 | -0.257 | 3.230 | 2.743 |
| ROUGE-L | -0.196 | -0.162 | 3.017 | 2.543 |
| METEOR  | -0.029 | -0.038 | 2.828 | 2.400 |
| **FEEL** | **0.404** | **0.314** | **2.049** | **1.657** |

### 4.3 Ablation Experiment

In order to further evaluate the effectiveness of mixed LLMs' emotional support capabilities, the ablation experiment of FEEL with part of LLMs is conducted. Table 4 shows the results of the combination of every two LLMs in FEEL on the sample.

First of all, compared with the single-LLM ablation model, the combination of the two models and FEEL achieve better results, which is reflected in the improvement of correlation and the reduction of both errors. Compared with the three ablated models combining two LLMs, FEEL achieves a notable improvement of 4.6% in Spearman's rank correlation coefficient against GLM+GPT. This enhancement is even more pronounced when FEEL is compared to ERNIE+GLM and ERNIE+GPT, with improvements of 63.56% and 53.03%, respectively, highlighting FEEL's robust performance in human alignments. Similarly, in Kendall's tau coefficient, FEEL shows significant improvements, particularly a 73.48% increase compared to ERNIE+GPT, demonstrating its strength in ordinal association evaluation. There's a slight decrease of 1.72% in Mean Absolute Error compared to GLM+GPT. Such a nuanced decrement may due to factors including measurement error, accidental differences in the data collection process, or the inherent complexity of the data itself.

Table 4. Ablation study results of FEEL.

|  | Spear. | Kend. | RMSE | MAE |
|---|---|---|---|---|
| ERNIE | 0.219 | 0.187 | 2.324 | 1.714 |
| GLM | 0.161 | 0.124 | 2.505 | 2.086 |
| GPT | 0.182 | 0.174 | 2.126 | 1.900 |
| ERNIE+GLM | 0.247 | 0.200 | 2.342 | 1.886 |
| ERNIE+GPT | 0.264 | 0.181 | 2.331 | 1.857 |
| GLM+GPT | 0.386 | 0.276 | 2.150 | **1.629** |
| **FEEL** | **0.404** | **0.314** | **2.049** | 1.657 |

## 5 Conclusion

In this paper, we introduce a LLMs-based evaluator FEEL to evaluate emotional support capability in dialogue systems. Meanwhile, we systematically and detailedly redefine the evaluation aspects of ESC and annotate a high-quality human score dataset ESCEval. Comparative experiment result shows that FEEL has higher human alignment than existing automatic evaluation metrics. Further ablation experiment shows that the scoring method that uses multiple LLMs for ensemble learning effectively improves the evaluation quality. However, although FEEL performs well in most aspects, it is still affected by noise including the subjectivity of manual scoring and differences in dialogue data. Our future work is to further enhance the robustness of FEEL. In addition, due to funding reasons, we only use three LLMs for model architecture in this paper. In the future, we will explore the impact of more LLM components on the evaluation effect.